\documentclass[runningheads]{llncs}

\usepackage{eccv}

\usepackage{eccvabbrv}

\usepackage{graphicx}
\usepackage{booktabs}
\usepackage{multirow}
\usepackage{arydshln}
\usepackage{soul} 
\usepackage{bm} 

\newcommand{\best}[1]{\bm{#1}}
\newcommand{\second}[1]{\underline{#1}}

\usepackage[table]{xcolor}

\usepackage[accsupp]{axessibility}  

\usepackage{hyperref}

\usepackage{orcidlink}
\usepackage{amsmath, amssymb}
\usepackage{array}
\usepackage{makecell}
\usepackage{notation}

\begin{document}

\title{CrossFeat: Bridging Imaging Modalities in Feature Descriptor Space}

\titlerunning{CrossFeat}

\author{Paul Schneider\inst{1,2}
\and
Nazim Haouchine\inst{1}
}

\authorrunning{P.~Schneider and N.~Haouchine}

\institute{Harvard Medical School, Brigham and Women’s Hospital, Boston, MA, USA \and
Technical University of Munich, Munich, Germany}

\maketitle

\begin{figure*}
    \centering
    \includegraphics[width=0.88\linewidth]{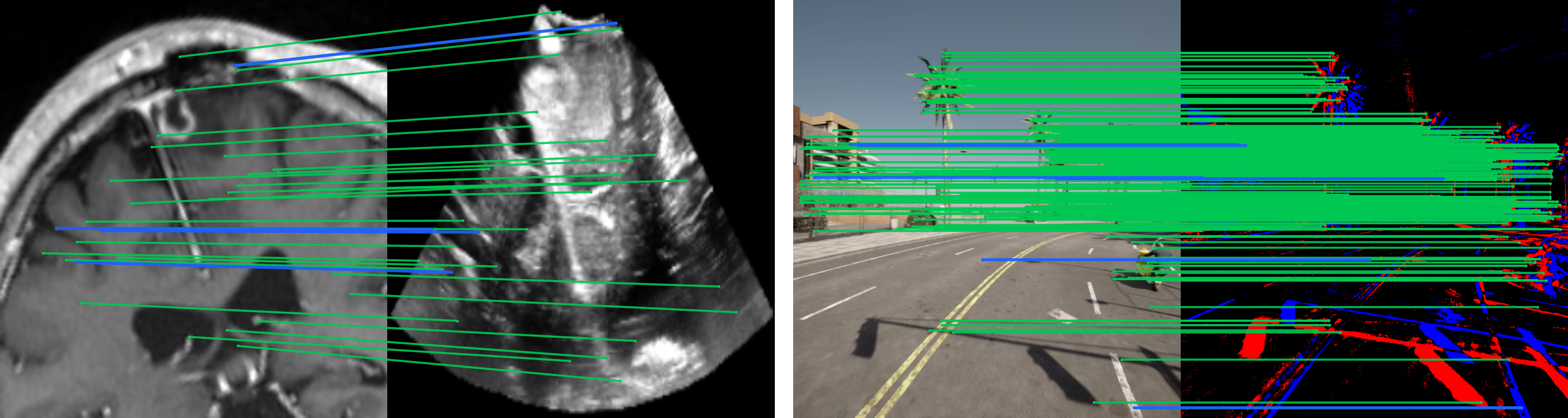}
    \caption{CrossFeat enables monomodal descriptors to operate across modalities. Examples on two challenging modality pairs: Crossing SuperPoint for MRI--Ultrasound images (left) and crossing SIFT for RGB--Event images (right). Blue lines show matches from the original descriptors, and green lines show \underline{increased} matches after crossing.}
\end{figure*}

\begin{abstract}
Most advances in keypoint descriptions address monomodal settings, where image variations arise from viewpoint, illumination, or contrast changes. 
Multimodal scenarios involve images produced by fundamentally different sensing processes, such as multispectral imaging, RGB-depth, satellite imagery, or medical imaging, causing the same structures to appear differently. 
A common solution to cross-modal description is to train descriptors for each modality pair, which requires retraining whenever the modalities change, or to train large models, which incur a significant increase in runtime.
Instead, we propose CrossFeat, a framework that enables an existing monomodal descriptor to operate across modalities. 
Our method learns a crossing function in descriptor space that maps features from one modality to a representation compatible with another. 
To preserve the structural information captured by the original descriptor, CrossFeat introduces a geometry–appearance disentanglement such that only appearance is altered while the geometric properties are preserved. 
Experiments across multiple domains and datasets demonstrate improved performance in multimodal matching. Code available here: \url{https://github.com/paulschneider01/CrossFeat}
\end{abstract}

\section{Introduction}

Keypoint matching is a fundamental computer vision task that underlies a wide range of applications, including registration, 3D reconstruction, object detection, retrieval, and tracking. A large body of work has been developed on this problem~\cite{liu2025modality}, achieving high accuracy and robustness, particularly with recent advances in learned descriptors. Most existing methods are designed for monomodal matching, reflecting the predominance of monomodal data, and introduce increasingly sophisticated mechanisms to improve robustness to geometric transformations as well as illumination and contrast variations. More recently, several approaches have proposed operating directly on already trained descriptors to compensate for specific limitations at test-time, such as enforcing rotational or affine invariance (e.g., steerers~\cite{steerers}) or accelerating matching~\cite{lindenberger2023lightglue}. These methods, however, primarily target geometric robustness or computational efficiency within a single modality. Multimodal matching can be viewed as a specialized subset of keypoint matching and arises in multiple settings~\cite{jiang2021review}, including multi-spectral scenarios such as near-infrared to visible-light matching~\cite{baruch2021joint}, visible-to-infrared~\cite{tuzcuouglu2024xoftr}, RGB-to-depth, and satellite imagery~\cite{jiang2021review}. In some cases, temporal variations~\cite{verdie2015tilde}, where the same scene is observed under substantially different conditions, can also be interpreted as a form of multimodality. Medical imaging poses additional challenges due to the inherently dynamic and heterogeneous nature of tissue appearance~\cite{heinrich2012mind,juvekar2023remind}. A common strategy to address multimodal matching is to supervise descriptor training on a specific pair of modalities~\cite{rasheed2024learning}; or to train large models~\cite{ren2025minima,he2025matchanything}, however, this approach requires retraining for each new modality pair, or which incur a significant increase in runtime.

In this paper, we propose \textbf{CrossFeat}, a novel framework for keypoint descriptors that enables an existing monomodal descriptor to "cross" modalities. We reframe multimodal matching as the learning of a non-linear operator that maps descriptors from one modality into a representation space compatible with another, without retraining or fine-tuning the original descriptor.
Our approach preserves the distinct characteristics and properties of the underlying descriptor, building upon decades of research in descriptor design rather than replacing it. CrossFeat can be applied to any chosen descriptor and extended to different modality pairs by training the crossing function for the specified descriptor and modalities, without modifying the original descriptor's training process. In summary, our main contributions are as follows.
\begin{itemize}
    \item We introduce a framework to learn a non-linear operator that translates any chosen descriptor to operate across at least two given modalities.
    \item We propose a disentanglement mechanism that separates a descriptor into appearance and geometry components, enabling the operator to act solely on the appearance component while retaining existing geometric properties.
    \item We conduct extensive evaluation across multiple datasets spanning medical, urban, and aerial domains, demonstrating improvements over state-of-the-art multimodal matching methods.
\end{itemize}
\section{Related Work}
\subsection{Multimodal Keypoint Descriptors and Matching}
Cross-spectral matching is among the earliest studied forms of multimodal correspondence in computer vision, emerging in surveillance, remote sensing, and biometrics where cross-spectral invariance is critical \cite{baruch2021joint,jiang2021review}. Early work adapted handcrafted descriptors through spectrum-specific normalization or joint embeddings, while recent approaches increasingly learn modality-agnostic correspondence functions. This shift reflects the diversity of modality gaps, which may arise from different sensing physics (e.g., thermal–visible, depth–RGB, MRI–ultrasound) or from temporal and seasonal changes that break photometric assumptions while preserving scene structure \cite{liu2025modality}.

The transition to learned detectors, descriptors, and matchers has been substantial. Context-aware assignment mechanisms, such as SuperPoint \cite{detone2018superpoint} coupled with graph-based attention matching \cite{sarlin2020superglue,lindenberger2023lightglue}, improved efficiency and scalability while maintaining robustness.
Detector-free matching further collapses detection, description, and matching into a single cross-attention procedure, exemplified by LoFTR \cite{sun2021loftr}. Cross-modal variants such as XoFTR extend this direction by adapting coarse-to-fine reasoning to visible–thermal pairs through explicit modality conditioning, improving robustness to spectral and illumination shifts \cite{tuzcuouglu2024xoftr}.
In medical imaging, modality gaps are compounded by anatomy-dependent visibility (bone, vessels, tumor, edema) and heterogeneous tissue appearance, motivating descriptors that encode shared structure or semantics rather than raw intensity \cite{heinrich2012mind,dey2024learning,dorent2023unified}. Early retinal fundus–FA/OCT registration relied on handcrafted descriptors \cite{chen2010partial}, 
but learning-based methods now dominate with deep supervised matching frameworks \cite{lee2019deep,rasheed2024learning}, semi-supervised and weakly supervised strategies \cite{liu2022semi,santarossa2022medregnet}, and  diffusion-based feature guidance \cite{tursynbek2025guiding}.

More recently, work has focused on scalability and generalization across modality pairs. MINIMA \cite{ren2025minima} trains on large synthetic multimodal datasets generated from RGB sources, achieving broad cross-modality performance without modality-specific modules. MatchAnything \cite{he2025matchanything} similarly relies on large-scale cross-modal pretraining to learn correspondences that generalize to unseen modality combinations. In parallel, foundation-model features are being adapted for correspondence; \cite{liu2025mind} analyzes the misalignment between single-image representation learning and cross-image matching and proposes strategies to improve feature matching capability.

\subsection{Descriptor Adaptation and Test-time Strategies}
A large body of work has explored improving matching performance without redesigning the full detection and description pipeline, by adapting descriptors, normalizing local frames, augmenting inference, or refining the matching stage.

Descriptor-space transforms operate directly on the feature representation at test time. Classical normalization strategies such as RootSIFT~\cite{rootsift} and VLAD~\cite{vlad} reshape descriptor distributions through power-law and intra-normalization to improve discriminability. Learning-based approaches including HardNet~\cite{hardnet} and SOSNet~\cite{sosnet} optimize the descriptor space with margin-based and second-order similarity objectives~\cite{hardnet,sosnet}, often combined with whitening. Equivariant formulations such as Steerers~\cite{steerers} and Affine Steerers~\cite{bokman2024affine} explicitly encode geometric transformations within the descriptor, increasing robustness to rotation and affine changes while maintaining compatibility with standard nearest-neighbor matching.

Patch-frame normalization instead adapts the local support region before descriptor extraction. AffNet learns affine-covariant regions to improve discriminability beyond repeatability~\cite{affnet}, while ASLFeat~\cite{aslfeat} and LIFT~\cite{lift} jointly optimize detection, localization, and description within unified frameworks. Polar Transformer Networks introduce canonicalization mechanisms that reduce geometric variability via explicit coordinate transformations~\cite{ptn}.

Inference-time augmentation leverages multiple forward passes or aggregation strategies to enhance robustness without additional training. R2D2 and D2-Net\cite{r2d2,d2net} jointly reason about detection reliability and descriptor distinctiveness. SuperPoint employs homography adaptation to aggregate predictions across synthetic transformations at inference~\cite{detone2018superpoint}, and Key.Net combines handcrafted and learned filters to improve stability under viewpoint changes~\cite{keynet}. 

Learned and adaptive matchers refine correspondences after descriptor extraction. Architectures such as SuperGlue~\cite{sarlin2020superglue}, LightGlue~\cite{lindenberger2023lightglue}, and LoFTR~\cite{sun2021loftr} incorporate contextual reasoning across keypoints to enforce geometric consistency and suppress outliers. Complementary approaches such as AdaLAM revisit handcrafted outlier filtering under adaptive constraints~\cite{adalam}, while ContextDesc augments local descriptors with broader contextual cues~\cite{contextdesc}. 

Collectively, these directions demonstrate that substantial gains can be achieved through descriptor adaptation and matching-time strategies, underscoring the flexibility of existing representations under carefully designed test-time transformations and contextual refinement.
However, a gap remains in bridging appearance differences induced by modality changes, particularly in adapting at test-time a previously trained descriptor to the specific modality against which it is being matched.

\section{Method}

\subsection{Problem Statement}
\label{sec:problem}

Let $\da, \db \in \mathbb{S}^{D-1}$ be $\ell_2$-normalized descriptors of size $D$ extracted at the same spatial location from modalities $\ma, \mb \in \modset$.
In practice, $\da$ can differ substantially from $\db$, as descriptor responses depend on modality-specific appearance.
We therefore seek to bridge this discrepancy by learning a \emph{crossing function}:
\begin{equation}
  \crosser : \mathbb{S}^{D-1} \times \modset \times \modset
    \to \mathbb{S}^{D-1}, \quad
  \dcrossed = \crosser(\da, \ma, \mb),
  \label{eq:crossing_fn}
\end{equation}
that transforms a source descriptor into the target feature space so that
$\dcrossed \approx \db$.
After crossing, standard nearest-neighbor matching in the target space yields
cross-modal correspondences.

\begin{figure*}[t]
    \centering
    \includegraphics[width=1\linewidth,
                     trim=0 12px 0 12px,
                     clip]{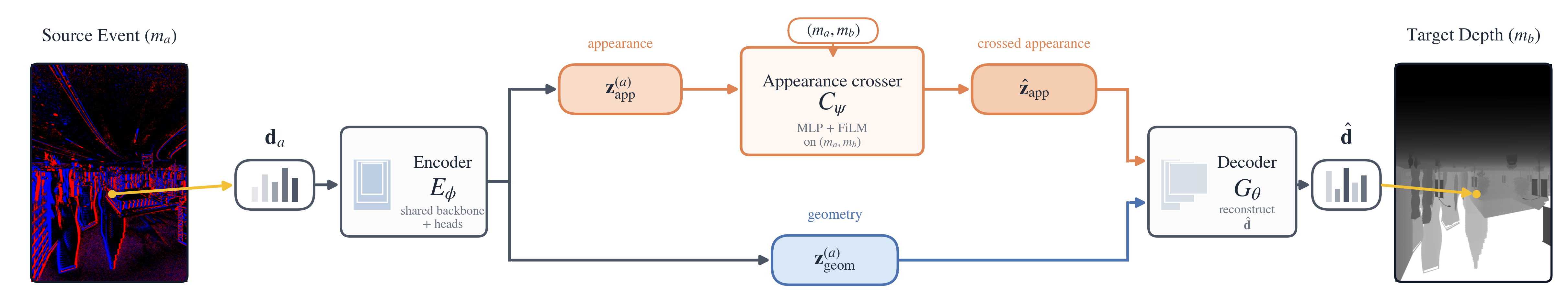}
    \caption{\textbf{Overview of CrossFeat.}
A descriptor $\mathbf{d}_a$ extracted from an image of modality $m_a$ is encoded by $E_{\phi}$ and decomposed into geometry $z_{\text{geom}}$ and appearance $z_{\text{app}}$. 
The appearance is transformed by the modality-conditioned crosser $C_{\psi}$ to obtain $\hat{z}_{\text{app}}$ compatible with the target modality $m_b$, while $z_{\text{geom}}$ is preserved. 
The decoder $G_{\theta}$ recombines $z_{\text{geom}}$ and $\hat{z}_{\text{app}}$ to reconstruct a descriptor $\hat{\mathbf{d}}$ in the target modality space, enabling cross-modal matching without retraining the original descriptor.}
    \label{fig:overview}
\end{figure*}

A descriptor encodes two types of information: (i)~\emph{geometry}, the local spatial structure (edges, corners, curvature) that is largely shared across modalities imaging the same scene, and (ii)~\emph{appearance}, modality-specific intensity and contrast patterns. Because structural cues may be attenuated or absent in some modalities (e.g., boundaries visible in MRI but not ultrasound), cross-modal crossing must transform appearance while preserving geometric content. We introduce an architecture that explicitly separates these components (Fig.~\ref{fig:overview}).

\subsection{Disentangled Descriptor Crossing}
\label{sec:vae_crosser}
\begin{figure*}[t]
    \centering
    \includegraphics[width=1.0\linewidth]{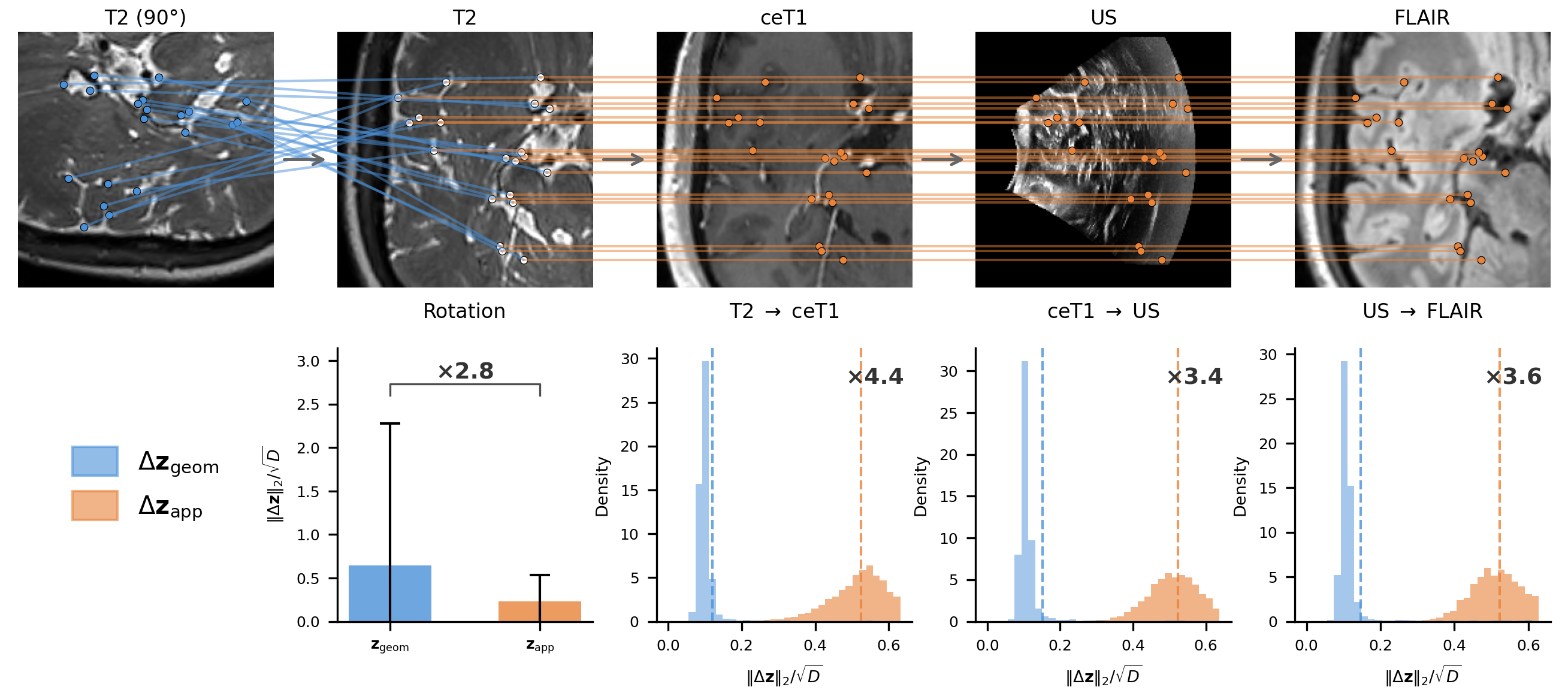}
    \caption{\textbf{Disentanglement of geometry and appearance.}
Top: correspondences across descriptors (T2, ceT1, US, FLAIR) obtained after crossing the appearance representation while preserving geometry. 
Bottom: sensitivity analysis of the latent components. Under rotation, the geometry component $z_{\text{geom}}$ changes significantly more than $z_{\text{app}}$. 
Conversely, across modality transitions, the appearance component exhibits larger variations than the geometry component, confirming that CrossFeat disentangles modality-dependent appearance from modality-invariant geometric structure.}
    \label{fig:disentanglement_proof}
\end{figure*}

A direct approach is to learn a single network
$\dcrossed = f_\psi(\da, \ma, \mb)$
that maps the full descriptor conditioned on the modality pair.
This treats the descriptor as a monolithic vector and relies on the network to
implicitly discover what to change and what to preserve.
In practice, we found that such models either alter geometry or under-transform appearance, ultimately hurting matching.
We therefore make the geometry--appearance decomposition architecturally explicit:
the descriptor is split into a modality-invariant \emph{geometry code} and a
modality-specific \emph{appearance code}, and crossing acts only on the latter (See Fig.~\ref{fig:disentanglement_proof}).

\vspace{0.5em}
\noindent\textbf{Encoder.}
A shared encoder $\encoder$ maps a descriptor from any modality to a geometry
code and an appearance code:
\begin{equation}
  \encoder : \mathbb{S}^{D-1} \to \mathbb{R}^{d_g} \times \mathbb{R}^{d_a},
  \quad
  \encoder(\mathbf{d}) = (\zgeom, \zapp).
  \label{eq:encoder}
\end{equation}
An MLP backbone produces an intermediate representation from which two
separate projection heads yield $\zgeom$ and $\zapp$.
Sharing the backbone across modalities forces the encoder to learn a common
representation space; the split into two heads is what enables selective
crossing.
Because $\zgeom$ bypasses the crosser, the architecture guarantees that geometry is unchanged during crossing. The training losses of Section~\ref{sec:optimization} provide the complementary pressure, encouraging $\zgeom$ to capture modality-invariant structure and $\zapp$ to absorb modality-specific statistics.
The appearance head is variational: it outputs a mean $\muapp$ and log-variance $\log \sigmaapp^2$, from which the appearance code is sampled via the reparameterization~\cite{kingma2013auto}:
\begin{equation}
  \zapp = \muapp + \sigmaapp \odot \boldsymbol{\varepsilon},
  \quad \boldsymbol{\varepsilon} \sim \mathcal{N}(\mathbf{0}, \mathbf{I}).
  \label{eq:reparam}
\end{equation}
The geometry code $\zgeom$ remains a deterministic point estimate. This asymmetry is deliberate: geometry must be a stable representation for reliable matching, whereas appearance benefits from stochastic regularization that prevents the encoder from memorizing modality-specific patterns. At inference, we set $\zapp = \muapp$ (no sampling).

\vspace{0.5em}
\noindent\textbf{Crosser.}
The crossing operation acts exclusively on the appearance code:
\begin{equation}
  \crosser : \mathbb{R}^{d_a} \times \modset \times \modset
    \to \mathbb{R}^{d_a}, \quad
  \zappcrossed
  = \crosser(\zapp^{(a)}, \ma, \mb).
  \label{eq:app_crosser}
\end{equation}
The crosser combines Feature-wise Linear Modulation (FiLM)~\cite{perez2018film} with a residual MLP. Learned modality embeddings for $\ma$ and $\mb$ are concatenated and passed through a small network that generates per-dimension affine parameters $(\gamma, \beta)$. The appearance code is first modulated by FiLM with $\gamma$ centered at identity for near-identity initialization: $\mathbf{z}_{\text{mod}}= (1 + \gamma) \odot \zapp^{(a)} + \beta$. A residual MLP then refines the modulated code, with a skip connection from the \emph{original} appearance:
\begin{equation}
  \zappcrossed = \zapp^{(a)} + \mathrm{MLP}(\mathbf{z}_{\text{mod}}).
  \label{eq:residual_crosser}
\end{equation}
The residual structure ensures that the crosser defaults to an identity mapping before training, and that gradients flow directly to the encoder.

\vspace{0.5em}
\noindent\textbf{Decoder.}
A decoder $\decoder$ reconstructs a descriptor from the two codes:
\begin{equation}
  \decoder : \mathbb{R}^{d_g} \times \mathbb{R}^{d_a}
    \to \mathbb{S}^{D-1}, \quad
  \hat{\mathbf{d}} = \decoder(\zgeom, \zapp).
  \label{eq:decoder}
\end{equation}
The decoder concatenates $\zgeom$ and $\zapp$, processes them through an MLP,
and $\ell_2$-normalizes the output to project it back onto~$\mathbb{S}^{D-1}$.
It receives FiLM conditioning on the target modality, allowing it to adapt
reconstruction to modality-specific descriptor statistics (e.g., different
gradient magnitude distributions across MRI sequences).

\vspace{0.5em}
\noindent\textbf{Full crossing path.}
At inference, given a source descriptor $\da$ from modality $\ma$, we encode
it, cross its appearance toward target modality~$\mb$, and decode:
\begin{equation}
  \dcrossed
    = \decoder\!\bigl(\zgeom^{(a)},\;
      \crosser(\zapp^{(a)}, \ma, \mb)\bigr).
  \label{eq:full_crossing}
\end{equation}
The crossed descriptor $\dcrossed$ lives in the feature space of~$\mb$ and can
be matched against target descriptors using standard nearest-neighbor search.

\subsection{Training Objective}
\label{sec:optimization}
The losses below serve three purposes: reconstruction of individual descriptors,
accurate crossing between modality spaces,
and disentanglement of geometry from appearance
in the latent codes.

\vspace{0.5em}
\noindent\textbf{Reconstruction loss.}
The latent codes must retain sufficient information to
recover the input.
For each modality we encode and decode the descriptor
and penalize the cosine distance to the original:
\[
  \Lrecon
  = \bigl(1 - \cosim(\decoder(\zgeom^{(a)}, \zapp^{(a)}),\, \da)\bigr)
  + \bigl(1 - \cosim(\decoder(\zgeom^{(b)}, \zapp^{(b)}),\, \db)\bigr).
\]
Without this term the encoder is free to discard
information, producing degenerate codes that satisfy
downstream objectives vacuously.\\
\textbf{Geometry alignment loss.}
Co-located descriptors from different modalities image
the same physical structure;
their geometry codes should therefore agree:
\[
  \Lgeom = \norm{\zgeom^{(a)} - \zgeom^{(b)}}_2^2.
  \label{eq:geom}
\]
The squared norm encourages soft alignment while
tolerating small differences arising from
modality-dependent visibility of structural features.
This is the primary disentanglement pressure:
it forces modality-specific information out of
$\zgeom$ and into $\zapp$.

\vspace{0.5em}
\noindent\textbf{Crossing loss.}
The primary supervision for descriptor crossing
measures how well a crossed descriptor reproduces
the target:
\begin{equation}
  \Lcross
  = 1 - \cosim\!\bigl(
      \decoder(\zgeom^{(a)},\,
        \crosser(\zapp^{(a)}, \ma, \mb)),\;
      \db\bigr).
  \label{eq:cross}
\end{equation}
The source geometry code $\zgeom^{(a)}$ is paired with
the crossed appearance code and decoded;
the result is compared to the actual target descriptor
$\db$ via cosine distance.
Note that geometry comes from the source, not the
target. $\crosser$ must encode appearance that, combined with source geometry alone, reconstructs
$\db$.

\vspace{0.5em}
\noindent\textbf{Adversarial modality loss.}
$\Lgeom$ encourages geometry agreement only at
paired points.
To enforce modality invariance globally, we attach
a linear modality classifier $p_\omega$ to $\zgeom$
and train it with gradient
reversal~\cite{ganin2016domain}:
\[
  \Ladv
  = -\mathbb{E}\bigl[
      \log p_\omega(m \mid \zgeom)\bigr],
\]
where gradients are negated before reaching the
encoder.
The classifier learns to predict modality from
$\zgeom$; gradient reversal makes the encoder actively
strip modality information, complementing the
pointwise constraint of $\Lgeom$.

\vspace{0.5em}
\noindent\textbf{Variational regularization.}
The variational appearance head (Eq.~\ref{eq:reparam}) is regularized by a KL divergence toward a learned, modality-conditioned prior~\cite{khemakhem2020variational}:
\[
  \Lkl = \KL\!\bigl(q(\zapp \mid \mathbf{d}) \;\|\;p(\zapp \mid m)\bigr),
\]
where $q(\zapp \mid \mathbf{d})  = \mathcal{N}(\muapp, \diag(\sigmaapp^2))$ is the encoder posterior and $p(\zapp \mid m) = \mathcal{N}(\boldsymbol{\mu}_m, \diag(\boldsymbol{\sigma}_m^2))$ is a prior whose mean and variance are learnable parameters indexed by modality~$m$. Conditioning the prior on the modality label makes this an identifiable Variational Autoencoder (VAE)~\cite{khemakhem2020variational}: the modality acts as an observed auxiliary variable that enables identification of the latent factors. A free-bits mechanism~\cite{kingma2016improved} prevents posterior collapse by exempting each latent dimension from the penalty when its per-dimension KL falls below a threshold. No KL penalty is applied to $\zgeom$, which remains deterministic.

\vspace{0.5em}
\noindent\textbf{Contrastive discrimination loss.}
$\Lcross$ encourages a crossed descriptor to be close
to its target but provides no incentive to be far from
other targets.
We add an InfoNCE loss~\cite{oord2018representation}
over crossed--target pairs within the batch:
\[
  \Lnce
  = -\frac{1}{B}\sum_{i=1}^{B}
    \log
    \frac{
      \exp\bigl(\cosim(\dcrossed_i,\, \mathbf{d}_{b,i}
)\,/\,\tau\bigr)
    }{
      \sum_{j=1}^{B}
      \exp\bigl(\cosim(\dcrossed_i,\, \mathbf{d}_{b,j})\,/\,\tau\bigr)
    },
\]

where $\tau$ is a temperature hyperparameter and $B$ the batch
size.
This provides hard-negative discrimination, which is
critical for nearest-neighbor matching where the
correct target must rank above all alternatives.

\vspace{0.5em}
\noindent\textbf{Full objective.}
The complete loss combines data-fidelity terms with disentanglement pressures and variational regularization:
\begin{equation}
  \Ltotal
  = \lambda_r \Lrecon
  + \lambda_g \Lgeom
  + \lambda_c \Lcross
  + \lambda_{\text{adv}} \Ladv
  + \lambda_{\text{kl}} \Lkl
  + \lambda_{\text{nce}} \Lnce.
  \label{eq:loss_overview}
\end{equation}
where $\lambda_r$, $\lambda_g$, $\lambda_c$, $\lambda_{\text{adv}}$, $\lambda_{\text{kl}}$, and $\lambda_{\text{nce}}$ are set to 1.0, 0.5, 1.0, 0.03, 0.01, and 0.5, respectively. These weights were selected once through an offline hyperparameter search over candidate configurations ($\approx 1000$ runs) and then kept fixed for all reported tasks and domains without retuning. Gradients of $\Ltotal$ update the shared encoder
$\encoder$, the appearance crosser $\crosser$, and
the decoder $\decoder$ jointly.
\renewcommand{\arraystretch}{0.85}
\setlength{\aboverulesep}{0.2pt}
\begin{table*}[t]
\centering
\caption{Evaluation results for sparse matching across tasks. All methods are capped at 500 matches per image.
Best results are \textbf{bold}, second best are \underline{underlined}.}
\label{tab:eval_results_sparse}
\small
\setlength{\tabcolsep}{4pt}

\resizebox{\textwidth}{!}{%
\begin{tabular}{l l ccccccccc}
\toprule
Task & Method & P & R & \#M & SR@1 & SR@3 & SR@5 & AUC@1 & AUC@3 & AUC@5 \\
\midrule

\multirow{8}{*}{Medical}
& SP+LG 
& 0.73 & 0.19 & 74 & 23.8 & 67.5 & 72.5 & 0.26 & 0.45 & 0.56 \\
& DISK+LG 
& 0.28 & 0.05 & 43 & 2.5 & 8.8 & 22.5 & 0.05 & 0.12 & 0.18 \\
& ALIKED+LG 
& 0.56 & 0.12 & 39 & 6.2 & 27.5 & 35.0 & 0.18 & 0.33 & 0.42 \\
& SIFT+LG 
& 0.39 & 0.06 & 34 & 2.5 & 15.0 & 25.0 & 0.08 & 0.19 & 0.27 \\
& SP+MINIMA-LG 
& 0.80 & \underline{0.26} & 91 & 37.5 & 67.5 & 82.5 & 0.26 & 0.49 & 0.60 \\
& SIFT+NN 
& 0.65 & 0.13 & 34 & 11.2 & 15.0 & 22.5 & 0.64 & 0.64 & 0.65 \\
& \textbf{Cross(SIFT)+NN (Ours)} 
& \underline{0.95} & 0.25 & 62 & \underline{47.0} & \underline{68.9} & \underline{88.4} & \underline{0.95} & \underline{0.95} & \underline{0.95} \\
& \textbf{Cross(SIFT)+NN+TTA (Ours)} 
& \textbf{0.98} & \textbf{0.44} & 107 & \textbf{79.2} & \textbf{97.4} & \textbf{97.4} & \textbf{0.96} & \textbf{0.96} & \textbf{0.97} \\

\midrule

\multirow{8}{*}{Driving}
& SP+LG 
& 0.29 & \underline{0.06} & 42 & 0.0 & 7.1 & 13.6 & 0.06 & 0.16 & 0.21 \\
& DISK+LG 
& 0.12 & 0.02 & 26 & 0.0 & 0.3 & 1.0 & 0.02 & 0.06 & 0.08 \\
& ALIKED+LG 
& 0.22 & 0.04 & 37 & 0.0 & 4.8 & 8.8 & 0.04 & 0.11 & 0.15 \\
& SIFT+LG 
& 0.16 & 0.01 & 11 & 0.3 & 2.7 & 5.3 & 0.04 & 0.09 & 0.11 \\
& SP+MINIMA-LG 
& 0.40 & 0.11 & 59 & 0.4 & 17.0 & 29.7 & 0.08 & 0.20 & 0.28 \\
& SIFT+NN 
& 0.41 & \underline{0.06} & 34 & 9.4 & 12.7 & 17.0 & 0.39 & 0.39 & 0.39 \\
& \textbf{Cross(SIFT)+NN (Ours)} 
& \underline{0.96} & 0.05 & 22 & \underline{73.5} & \underline{74.9} & \underline{75.5} & \underline{0.92} & \underline{0.92} & \underline{0.92} \\
& \textbf{Cross(SIFT)+NN+TTA (Ours)} 
& \textbf{0.97} & \textbf{0.23} & 110 & \textbf{85.3} & \textbf{88.6} & \textbf{90.7} & \textbf{0.94} & \textbf{0.94} & \textbf{0.94} \\

\midrule

\multirow{8}{*}{Satellite}
& SP+LG 
& 0.00 & 0.00 & 28 & 0.0 & 0.0 & 0.0 & 0.00 & 0.00 & 0.00 \\
& DISK+LG 
& 0.00 & 0.00 & 30 & 0.0 & 0.0 & 0.0 & 0.00 & 0.00 & 0.00 \\
& ALIKED+LG 
& 0.01 & 0.00 & 22 & 0.0 & 0.0 & 0.0 & 0.00 & 0.00 & 0.00 \\
& SIFT+LG 
& 0.00 & 0.00 & 16 & 0.0 & 0.0 & 0.0 & 0.00 & 0.00 & 0.00 \\
& SP+MINIMA-LG 
& 0.01 & 0.00 & 9 & 0.0 & 0.0 & 0.0 & 0.00 & 0.00 & 0.01 \\
& SIFT+NN 
& 0.23 & 0.02 & 26 & 0.0 & 0.0 & 0.0 & 0.23 & 0.23 & 0.23 \\
& \textbf{Cross(SIFT)+NN (Ours)} 
& \textbf{0.87} & \underline{0.02} & 9 & \textbf{34.0} & \textbf{38.0} & \textbf{42.0} & \textbf{0.87} & \textbf{0.87} & \textbf{0.87} \\
& \textbf{Cross(SIFT)+NN+TTA (Ours)} 
& \underline{0.74} & \textbf{0.05} & 20 & \underline{18.0} & \underline{23.0} & \underline{29.0} & \underline{0.73} & \underline{0.73} & \underline{0.73} \\

\bottomrule
\end{tabular}}
\end{table*}

\section{Experiments}

\subsection{Datasets, Metrics and Training}
\noindent\textbf{Datasets and Tasks.}
We evaluate our method across three multimodal matching tasks spanning medical imaging, autonomous driving, and satellite imagery: 
(i) a Medical task, we train on the ReMIND dataset~\cite{juvekar2023remind} composed of brain images acquired using ceT1, T2, FLAIR, and ultrasound modalities, and evaluate on BRATS (ceT1--T2, ceT1--FLAIR)~\cite{menze2014multimodal} and RESECT (T1--US)~\cite{xiao2017re}, 
(ii) a Driving task, where we train on the synthetic EventScape dataset~\cite{gehrig2021combining} (RGB, depth, event) or urban images and evaluate on the real-world DELIVER dataset~\cite{zhang2023delivering},
and (iii) a Satellite task, where we train on WHU-OPT-SAR (RGB, SAR)~\cite{li2022mcanet} of satellite images and evaluate on QXS-SAROPT~\cite{huang2021qxs}. This yields 10 modality pairs across tasks; the architecture scales to arbitrary $N$.

\vspace{0.5em}
\noindent\textbf{Evaluation Metrics.}
We evaluate CrossFeat against dense matching methods using Success Rate (SR) and the Area Under the Curve (AUC) at different thresholds. In addition, when comparing with sparse methods we also report Precision, Recall, and the average number of matches per image.

\vspace{0.5em}
\noindent\textbf{Training and Implementation details.}
We extract descriptors at keypoint locations detected independently in each modality. To ensure balanced spatial coverage, we adopt a dual-anchor sampling strategy: half of the training keypoints are detected in the source modality and half in the target modality. This avoids biasing the training distribution toward structures salient in only one modality. Descriptors from all modalities are pooled and whitened via PCA. 
We train one crossing model per task, each handling all modality pairs within its domain. The crossing network (Sec.~\ref{sec:vae_crosser}) uses latent dimensions $d_g{=}D$ and $d_a{=}D/2$ with hidden dimension $2D$, totaling ${\sim}0.5$--$0.6$M parameters. 
Training takes ${\sim}$1--3 hours on an NVIDIA A100 GPU. 
Images are contrast-normalized with CLAHE~\cite{zuiderveld1994contrast} before descriptor extraction. We train with AdamW~\cite{loshchilov2019decoupled} ($\beta_1{=}0.9$, $\beta_2{=}0.999$, weight decay $10^{-4}$) and cosine annealing, using batch size 2048. The learning rate is $3{\times}10^{-4}$ for all tasks. Training runs up to 500 epochs, with early stopping on validation top-1 accuracy and 40 epochs patience. At test time, we optionally apply a lightweight adaptation step (TTA) after crossing. After an AdaIN warm-up, we run five iterations in which source and target descriptors are matched by mutual nearest neighbors, geometrically consistent pairs are selected by rigid RANSAC, and these inliers are used as pseudo-labels to fit a closed-form ridge regression predicting a small residual correction for the crossed source descriptors. The corrected descriptors are added residually and $\ell_2$-normalized before the next iteration. Because TTA depends on the model's own RANSAC inliers, it improves alignment when initial matches are reliable, but can reduce precision when they are too few or noisy. Dense end-to-end matchers do not expose an equivalent per-keypoint descriptor interface, and therefore this TTA procedure is only applied to sparse descriptor-based methods. Additional results are provided in the supplementary material.

\renewcommand{\arraystretch}{0.85}
\setlength{\aboverulesep}{0.2pt}
\begin{table*}[t]
\centering
\caption{Evaluation results for dense matching across tasks. All methods are capped at 500 matches per image.
Best results are \textbf{bold}, second best are \underline{underlined}.}
\label{tab:eval_results_dense}
\small
\setlength{\tabcolsep}{4pt}

\resizebox{\textwidth}{!}{%
\begin{tabular}{l l cccccc}
\toprule
Task & Method & SR@1 & SR@3 & SR@5 & AUC@1 & AUC@3 & AUC@5 \\
\midrule

\multirow{5}{*}{Medical}
& MatchAnything 
& 3.8 & 6.2 & 17.5 & 0.23 & 0.28 & 0.33 \\
& MINIMA-LoFTR 
& 17.5 & 85.0 & \underline{97.5} & 0.19 & 0.57 & 0.70 \\
& MINIMA-RoMA 
& \underline{66.2} & \textbf{97.5} & \textbf{100.0} & 0.46 & 0.74 & 0.83 \\
& \textbf{Cross(SIFT)+NN (Ours)} 
& 47.0 & 68.9 & 88.4 & \underline{0.95} & \underline{0.95} & \underline{0.95} \\
& \textbf{Cross(SIFT)+NN+TTA (Ours)} 
& \textbf{79.2} & \underline{97.4} & 97.4 & \textbf{0.96} & \textbf{0.96} & \textbf{0.97} \\

\midrule

\multirow{5}{*}{Driving}
& MatchAnything 
& 7.0 & 10.3 & 15.0 & 0.54 & 0.55 & 0.56 \\
& MINIMA-LoFTR 
& 32.3 & 68.3 & 74.0 & 0.40 & 0.60 & 0.66 \\
& MINIMA-RoMA 
& 3.3 & 34.3 & 61.0 & 0.15 & 0.41 & 0.53 \\
& \textbf{Cross(SIFT)+NN (Ours)} 
& \underline{73.5} & \underline{74.9} & \underline{75.5} & \underline{0.92} & \underline{0.92} & \underline{0.92} \\
& \textbf{Cross(SIFT)+NN+TTA (Ours)} 
& \textbf{85.3} & \textbf{88.6} & \textbf{90.7} & \textbf{0.94} & \textbf{0.94} & \textbf{0.94} \\

\midrule

\multirow{5}{*}{Satellite}
& MatchAnything 
& 0.0 & 0.0 & 0.0 & 0.00 & 0.00 & 0.00 \\
& MINIMA-LoFTR 
& 1.0 & 1.0 & 2.0 & 0.01 & 0.02 & 0.03 \\
& MINIMA-RoMA 
& 0.0 & 0.0 & 1.0 & 0.00 & 0.01 & 0.02 \\
& \textbf{Cross(SIFT)+NN (Ours)} 
& \textbf{34.0} & \textbf{38.0} & \textbf{42.0} & \textbf{0.87} & \textbf{0.87} & \textbf{0.87} \\
& \textbf{Cross(SIFT)+NN+TTA (Ours)} 
& \underline{18.0} & \underline{23.0} & \underline{29.0} & \underline{0.73} & \underline{0.73} & \underline{0.73} \\

\bottomrule
\end{tabular}}
\end{table*}

\begin{figure*}
   \centering
    \includegraphics[width=1.0\linewidth, trim=3cm 2cm 3cm 1cm, clip]{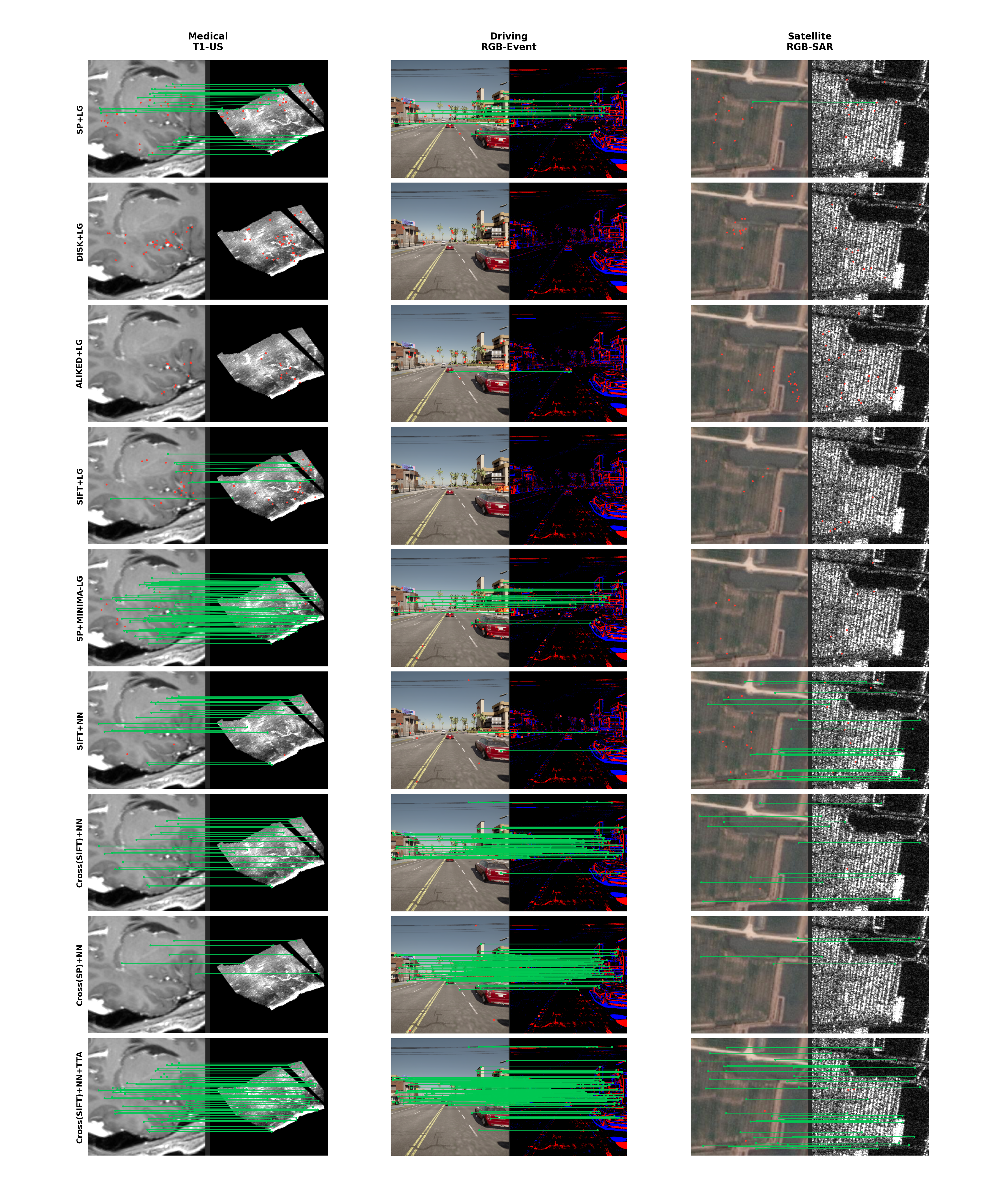}
    \caption{Qualitative multimodal matching results:
Example correspondences across medical task (T1–US), driving task (RGB–Event), and satellite task (RGB–SAR). 
Applying CrossFeat to existing descriptors (bottom rows) increases the number of correct matches (green lines), and reduce the number of outliers (red dots) across modalities w.r.t related works, demonstrating improved cross-modal compatibility without retraining the original descriptor.}
\label{fig:sparse_grid}
\end{figure*}

\subsection{Comparison with state-of-the-art methods}
\noindent\textbf{Sparse Methods.} We first compare CrossFeat against a set of representative sparse matching pipelines combining different descriptors and matchers, including SuperPoint (SP)~\cite{detone2018superpoint}, DISK~\cite{tyszkiewicz2020disk}, ALIKED~\cite{zhao2023aliked}, and SIFT~\cite{lowe2004distinctive} paired with LightGlue (LG)~\cite{lindenberger2023lightglue}, as well as SuperPoint paired with MINIMA-LG~\cite{ren2025minima} and SIFT with nearest-neighbor matching (NN). Table~\ref{tab:eval_results_sparse} and Figure~\ref{fig:sparse_grid} reports results across the three evaluation tasks. On the Medical task, CrossFeat, applied on SIFT, improves performance across most metrics, achieving the best results in SR and AUC. On the Driving task, baseline methods obtain very low success rates, while CrossFeat substantially increases matching accuracy, again achieving the best SR and AUC values. On the Satellite task, which is particularly challenging, most baselines fail to produce reliable matches, whereas CrossFeat remains the only method achieving non-zero success rates across thresholds, with the strongest performance obtained without TTA. These results demonstrate consistent improvements across domains and modality pairs. 

CrossFeat primarily improves match reliability rather than match density. It often retrieves fewer correspondences, leading to comparatively lower recall, but the retained matches are more geometrically consistent. This explains the strong SR@1 and AUC@1 values despite modest recall. TTA increases recall by recovering additional matches, but can reduce precision when the initial inlier set is too small or noisy, as in the Satellite task.

\vspace{0.5em}
\noindent\textbf{Dense Methods.} We further compare CrossFeat with recent dense matching methods, including MatchAnything~\cite{he2025matchanything}, MINIMA-LoFTR~\cite{sun2021loftr,ren2025minima}, and MINIMA-RoMa~\cite{edstedt2024roma,ren2025minima}. Table~\ref{tab:eval_results_dense} reports SR and AUC metrics across the three tasks. On the medical task, CrossFeat with TTA achieves the highest AUC across all thresholds and the best SR@1, while MINIMA-RoMa attains the highest SR at larger thresholds. On the driving task, CrossFeat substantially outperforms all baselines, achieving the highest SR and AUC values across thresholds. On the satellite task, dense methods struggle to produce reliable correspondences, whereas CrossFeat remains the only approach achieving consistent success rates and high AUC values.

As for \textbf{runtime}, our method takes $\sim$0.5--1\,s per image to perform detection, description, crossing, and matching (can be sped up when using LightGlue instead of NN), while dense matching methods require roughly 5\,s per image.

\begin{figure*}[t]
   \centering
    \includegraphics[width=1\linewidth]{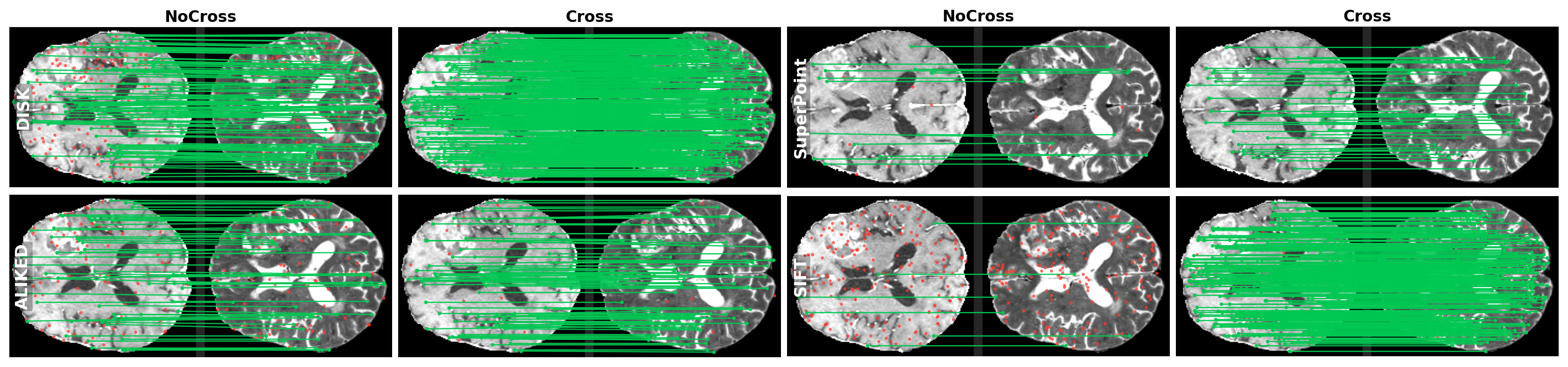}
    \caption{Examples of matching between ceT1 and T2 modalities. Applying CrossFeat on SIFT, ALIKED, SuperPoint, and DISK consistently improves their performances.}
        \label{fig:boost_quali}
\end{figure*}

\subsection{CrossFeat improves descriptor performances}
\begin{figure*}[t]
\centering
\includegraphics[width=1.0\linewidth]{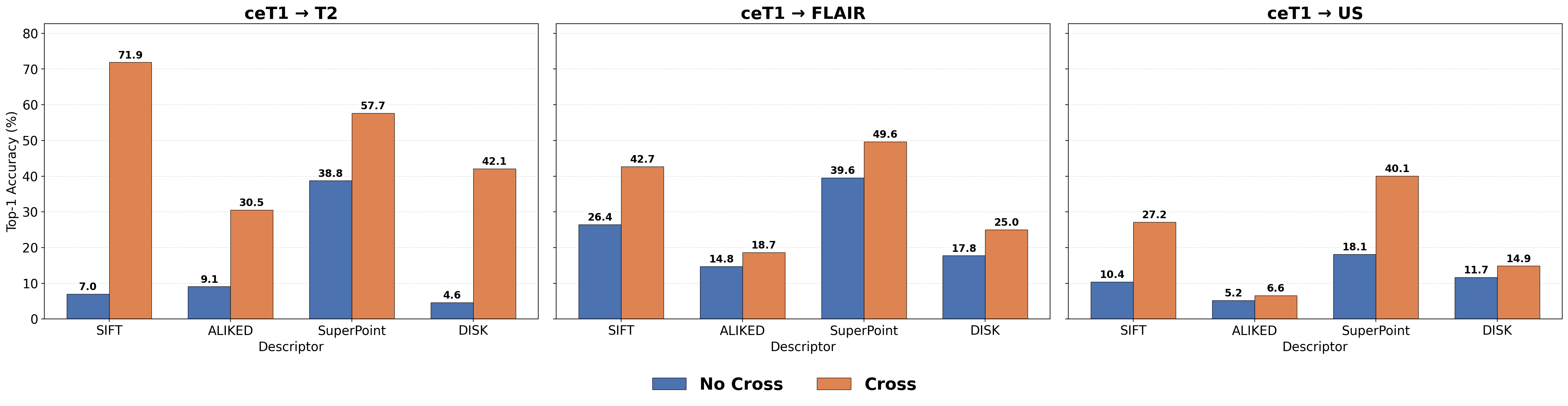}
\caption{Top-1 accuracy for ceT1 matched to T2, US, and FLAIR. CrossFeat consistently improves the descriptors SIFT, ALIKED, SuperPoint, and DISK.}
\label{fig:boost_test}
\end{figure*}

We perform experiments across multiple medical modality pairs (ceT1--T2, ceT1--FLAIR, T1--US) to analyze how CrossFeat improves descriptor performance. 
As reported in Figure ~\ref{fig:boost_test}, applying CrossFeat consistently improves all evaluated descriptors, including SIFT, SuperPoint, ALIKED, and DISK. The gains are particularly pronounced for SIFT, but also observed for learned descriptors, demonstrating that CrossFeat can enhance both classical and learned representations by adapting them to multimodal appearance changes. Figure~\ref{fig:boost_quali} illustrates the matching improvement on one sample.

\begin{figure*}
    \centering
    \includegraphics[width=0.9\linewidth]{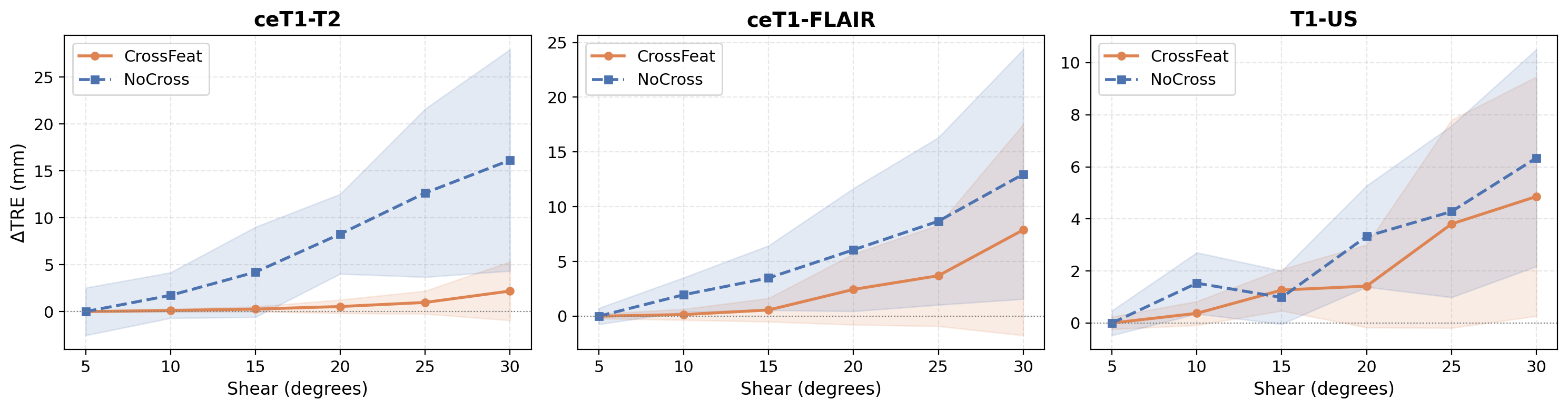}
    \caption{Robustness to geometric changes under increasing shear for three modality pairs. 
CrossFeat consistently yields lower error ($\Delta$TRE) than the baseline (NoCross).}
    \label{fig:boost_shear}
\end{figure*}

Using CrossFeat improves robustness to geometric distortions. As shown in Fig.~\ref{fig:boost_shear}, the error ($\Delta$TRE) increases much more slowly under increasing shear when using CrossFeat compared to the baseline without crossing, indicating that crossing the descriptors preserves geometric properties of the original descriptor while improving tolerance to appearance changes.

\begin{figure*}
   \centering
    \includegraphics[width=1.0\linewidth]{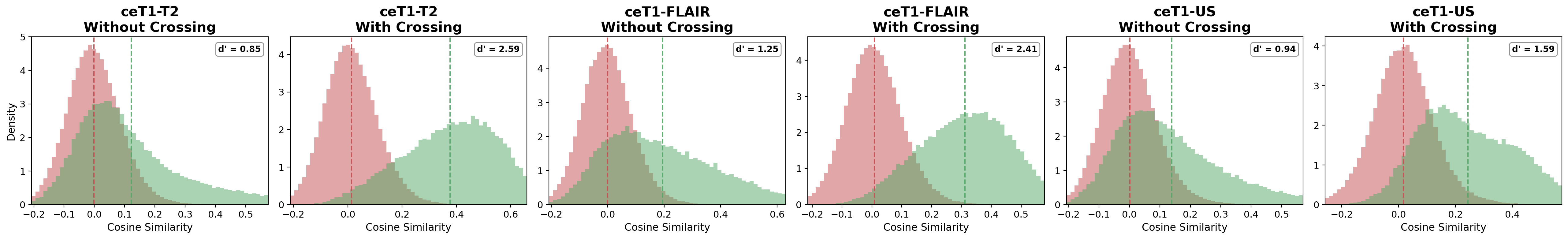}
    \caption{Cosine similarity distributions for matching (green) and non-matching (red) keypoints. Crossing the appearance representation increases the separation between the two distributions, yielding higher sensitive values $d'$ across modality pairs.}
        \label{fig:separation}
\end{figure*}

We analyze descriptor discriminability by measuring the cosine similarity between descriptors of matching and non-matching keypoints across modality pairs (Fig.~\ref{fig:separation}). Crossing the descriptor shifts the similarity of true correspondences toward higher values while leaving non-matching pairs largely unchanged. This increases the separation between the two distributions and leads to significantly higher sensitive values $d'$, indicating that CrossFeat improves the ability of the descriptor to distinguish correct matches from distractors in multimodal settings.

\definecolor{lightgray}{gray}{0.92}
\renewcommand{\arraystretch}{0.85}
\setlength{\aboverulesep}{0.2pt}
\begin{table*}[t]
\centering
\caption{Ablation study on the ReMIND dataset. Each row removes one component from the full model.
Best results are \textbf{bold}, second best are \underline{underlined}.} 
\label{tab:ablation}
\small
\setlength{\tabcolsep}{4pt}

\resizebox{\textwidth}{!}{%
\begin{tabular}{l ccccccc}
\toprule
\textbf{Configuration} & P & R & \#M & SR@1 & SR@3 & AUC@1 & AUC@3 \\
\midrule
\addlinespace[6pt]

Full model
& $\best{91.7{\scriptstyle\pm3.4}}$
& $\second{6.2{\scriptstyle\pm0.3}}$
& $\second{91.2{\scriptstyle\pm2.1}}$
& $\best{52.9{\scriptstyle\pm4.8}}$
& $\second{65.1{\scriptstyle\pm10.3}}$
& $\best{89.3{\scriptstyle\pm6.0}}$
& $\best{89.3{\scriptstyle\pm6.0}}$
\\

\addlinespace[6pt]

\multicolumn{8}{l}{\textbf{Architecture}} \\
\midrule

w/o disentanglement
& $84.8{\scriptstyle\pm7.6}$
& $5.9{\scriptstyle\pm0.2}$
& $86.8{\scriptstyle\pm2.1}$
& $39.6{\scriptstyle\pm7.2}$
& $60.0{\scriptstyle\pm7.0}$
& $80.8{\scriptstyle\pm4.5}$
& $80.8{\scriptstyle\pm4.5}$
\\

\rowcolor{lightgray}
w/o variational
& $86.3{\scriptstyle\pm5.8}$
& $\best{6.4{\scriptstyle\pm0.3}}$
& $\best{93.2{\scriptstyle\pm4.7}}$
& $48.3{\scriptstyle\pm6.4}$
& $65.0{\scriptstyle\pm4.4}$
& $84.2{\scriptstyle\pm3.8}$
& $84.2{\scriptstyle\pm3.8}$
\\

w/o adversarial
& $84.1{\scriptstyle\pm7.9}$
& $5.8{\scriptstyle\pm0.2}$
& $84.9{\scriptstyle\pm4.6}$
& $40.6{\scriptstyle\pm10.3}$
& $58.7{\scriptstyle\pm11.7}$
& $84.0{\scriptstyle\pm7.9}$
& $84.0{\scriptstyle\pm7.9}$
\\

\rowcolor{lightgray}
w/o cond.\ decoder
& $\second{88.4{\scriptstyle\pm9.8}}$
& $\second{6.2{\scriptstyle\pm0.4}}$
& $88.7{\scriptstyle\pm5.2}$
& $40.9{\scriptstyle\pm1.7}$
& $60.7{\scriptstyle\pm6.1}$
& $82.4{\scriptstyle\pm4.9}$
& $82.4{\scriptstyle\pm4.9}$
\\

\addlinespace[6pt]
\multicolumn{8}{l}{\textbf{Losses}} \\
\midrule

w/o $\mathcal{L}_{\mathrm{ctr}}$
& $75.0{\scriptstyle\pm5.5}$
& $3.2{\scriptstyle\pm0.1}$
& $43.9{\scriptstyle\pm1.6}$
& $15.1{\scriptstyle\pm5.9}$
& $28.1{\scriptstyle\pm7.2}$
& $73.4{\scriptstyle\pm5.1}$
& $73.9{\scriptstyle\pm5.5}$
\\

\rowcolor{lightgray}
w/o $\mathcal{L}_{\mathrm{geom}}$
& $84.4{\scriptstyle\pm8.8}$
& $6.0{\scriptstyle\pm0.2}$
& $86.6{\scriptstyle\pm2.8}$
& $\second{48.9{\scriptstyle\pm10.6}}$
& $\best{65.3{\scriptstyle\pm6.2}}$
& $82.2{\scriptstyle\pm6.9}$
& $82.2{\scriptstyle\pm6.9}$
\\

w/o $\mathcal{L}_{\mathrm{recon}}$
& $86.3{\scriptstyle\pm6.2}$
& $5.5{\scriptstyle\pm0.3}$
& $77.3{\scriptstyle\pm3.2}$
& $44.4{\scriptstyle\pm5.3}$
& $57.9{\scriptstyle\pm5.7}$
& $\second{84.3{\scriptstyle\pm3.7}}$
& $\second{84.3{\scriptstyle\pm3.7}}$
\\

\rowcolor{lightgray}
w/o latent reg.
& $82.8{\scriptstyle\pm8.2}$
& $\second{6.2{\scriptstyle\pm0.2}}$
& $88.2{\scriptstyle\pm4.4}$
& $41.0{\scriptstyle\pm7.2}$
& $60.9{\scriptstyle\pm5.9}$
& $78.7{\scriptstyle\pm4.1}$
& $78.7{\scriptstyle\pm4.1}$
\\

\addlinespace[6pt]
\multicolumn{8}{l}{\textbf{Training}} \\
\midrule

w/o KP sampling
& $69.5{\scriptstyle\pm2.5}$
& $1.0{\scriptstyle\pm0.1}$
& $11.2{\scriptstyle\pm0.8}$
& $45.4{\scriptstyle\pm6.6}$
& $60.3{\scriptstyle\pm4.2}$
& $61.5{\scriptstyle\pm2.1}$
& $61.7{\scriptstyle\pm2.1}$
\\

\addlinespace[6pt]
\bottomrule
\end{tabular}}
\end{table*}

\subsection{Ablation study}
Table~\ref{tab:ablation} reports ablation results on the ReMIND dataset (held-out test set), averaged across six modality pairs and five random seeds. Since the loss weights are kept fixed across all experiments, this ablation analyzes the effect of each objective component by removing individual terms rather than retuning $\lambda$ values per task.

Removing the geometry-appearance disentanglement ($-13.3$\,pp SR@1) causes the largest drop, confirming that crossing only the appearance code while preserving geometry is essential. The adversarial discriminator ($-12.3$\,pp) and FiLM-conditioned decoder ($-12.0$\,pp) both contribute substantially, showing that the modality-invariance in $z_\text{geom}$ requires explicit enforcement and that modality-aware reconstruction improves descriptor quality. The variational bottleneck has a moderate effect ($-4.6$\,pp) while slightly increasing match count, suggesting it primarily acts as a regularizer.

The contrastive loss is the single most important component: removing it collapses SR@1 from $52.9\%$ to $15.1\%$ and halves match count, confirming that hard-negative discrimination is critical for nearest-neighbor matching. Reconstruction ($-8.5$\,pp) and latent regularization ($-10.6$\,pp AUC@3) both contribute meaningfully, while $\mathcal{L}_\text{geom}$ alone has limited impact ($-4.0$\,pp), likely because the adversarial loss provides a complementary invariance signal.

Replacing dual-anchor keypoint sampling with random sampling reduces matches from $89$ to $11$ and precision by $22$\,pp, confirming that training on structurally informative locations in both source and target modality is critical.

\section{Discussion and Conclusion}

The experiments show that CrossFeat enables standard monomodal descriptors to operate effectively in multimodal settings by adapting appearance while preserving the geometric structure encoded by the original representation. Consistent improvements are observed across different descriptors and modality pairs (see Fig~\ref{fig:boost_quali} and ~\ref{fig:boost_test}). Robustness experiments further show that crossing maintains geometric consistency under distortions (see Fig~\ref{fig:boost_shear}), while the similarity analysis demonstrates increased separation between matching and non-matching descriptors, leading to improved discriminability in multimodal conditions (see Fig~\ref{fig:separation}).

Across all tasks, CrossFeat consistently outperforms sparse matching pipelines (See Tab.~\ref{tab:eval_results_sparse}), particularly in challenging modality gaps where baseline methods often fail to produce reliable correspondences. Despite operating within the classical detector–descriptor–matching pipeline, CrossFeat achieves performance that is competitive with, and in several cases superior to, recent dense matching approaches (See Tab.~\ref{tab:eval_results_dense}). 
Importantly, CrossFeat is substantially more efficient than dense matchers during both training and inference, since it operates on sparse descriptors and relies on a lightweight crossing network rather than large correlation-based architectures, resulting in $\sim$100$\times$ fewer parameters and $\sim$5-10$\times$ faster inference.

A limitation of the current formulation is that the modality pair must be specified at inference time, as the crossing network learns modality-dependent transformations. Future work will explore estimating the modality gap directly from descriptors or image statistics and using this signal to guide the crossing process. Such a mechanism could adaptively determine the required appearance transformation and enable more flexible operation when modalities are unknown or vary continuously.

Overall, these results suggest that bridging the modality gap at the descriptor level is an effective strategy for multimodal correspondence. By enabling existing descriptors to generalize across sensing modalities, CrossFeat offers a simple and efficient alternative to training modality-specific descriptors or adopting entirely new matching architectures.

\section*{Acknowledgements}
The work was supported by the NIH Grants K25EB035166 and R03EB033910.

%
%
\bibliographystyle{splncs04}
\bibliography{main}
\end{document}